\documentclass[10pt,conference]{IEEEtran}
\IEEEoverridecommandlockouts

\usepackage{cite}
\usepackage{amsmath,amssymb,amsfonts}
\usepackage{algorithmic}
\usepackage{graphicx}
\usepackage{textcomp}
\usepackage{xcolor}
\usepackage{numprint}
\usepackage{siunitx}
\def\BibTeX{{\rm B\kern-.05em{\sc i\kern-.025em b}\kern-.08em
    T\kern-.1667em\lower.7ex\hbox{E}\kern-.125emX}}

\begin{document}

\title{Quantum Ensembling Methods for Healthcare and Life Science 
\thanks{This manuscript has been authored in part by UT-Battelle, LLC, under Contract No. DE-AC0500OR22725 with the U.S. Department of Energy. The United States Government retains and the publisher, by accepting the article for publication, acknowledges that the United States Government retains a non-exclusive, paid-up, irrevocable, world-wide license to publish or reproduce the published form of this manuscript, or allow others to do so, for the United States Government purposes. The Department of Energy will provide public access to these results of federally sponsored research in accordance with the DOE Public Access Plan.}
}

\author{
\IEEEauthorblockN{Kahn Rhrissorrakrai}
\IEEEauthorblockA{
\textit{IBM Research}\\
Yorktown Heights,NY, USA \\
krhriss@us.ibm.com}
\and
\IEEEauthorblockN{Kathleen E. Hamilton}
\IEEEauthorblockA{\textit{Computational Science and Engineering Division} \\
\textit{Oak Ridge National Laboratory}\\
Oak Ridge, TN USA \\
hamiltonke@ornl.gov}
\and
\IEEEauthorblockN{Prerana Bangalore Parthasarathy}
\IEEEauthorblockA{\textit{Lerner Research Institute} \\
\textit{Cleveland Clinic}\\
Cleveland, Ohio, USA \\
bangalap2@ccf.org}
\and
\IEEEauthorblockN{Aldo Guzm\'an-S\'aenz}
\IEEEauthorblockA{
\textit{IBM Research}\\
Yorktown Heights,NY, USA \\
aldo.guzman.saenz@ibm.com}
\and
\IEEEauthorblockN{Tyler Alban}
\IEEEauthorblockA{\textit{Lerner Research Institute} \\
\textit{Cleveland Clinic}\\
Cleveland, Ohio, USA \\
albant@ccf.org}
\and
\IEEEauthorblockN{Filippo Utro}
\IEEEauthorblockA{
\textit{IBM Research}\\
Yorktown Heights,NY, USA \\
futro@us.ibm.com}
\and
\IEEEauthorblockN{Laxmi Parida}
\IEEEauthorblockA{
\textit{IBM Research}\\
Yorktown Heights, NY, USA \\
parida@us.ibm.com}
}

\maketitle

\begin{abstract}
Learning on small data is a challenge frequently encountered in many real-world applications.  In this work we study how effective quantum ensemble models are when trained on small data problems in healthcare and life sciences.  We constructed multiple types of quantum ensembles for binary classification using up to 26 qubits in simulation and 56 qubits on quantum hardware.  Our ensemble designs use minimal trainable parameters but require long-range connections between qubits. We tested these quantum ensembles on synthetic datasets and gene expression data from renal cell carcinoma patients with the task of predicting patient response to immunotherapy. From the performance observed in simulation and initial hardware experiments, we demonstrate how quantum embedding structure affects performance and discuss how to extract informative features and build models that can learn and generalize effectively.  We present these exploratory results in order to assist other researchers in the design of effective learning on small data using ensembles. Incorporating quantum computing in these data constrained problems offers hope for a wide range of studies in healthcare and life sciences where biological samples are relatively scarce given the feature space to be explored.

\end{abstract}


\section{Introduction}
Artificial intelligence and machine learning (AI/ML) have led to incredible advancements in healthcare and life science (HCLS).  However, biological and healthcare data poses many challenges for AI algorithms, including complexity and scale challenges \cite{RN899}, learning in sample-limited scenarios, model overfitting, saturated learning \cite{RN901,RN1047}, quadratic time and space complexity, and poor generalization. Moreover, there has been significant progress in quantum computing technologies that may offer the opportunity of overcoming some of those limitations in specific use cases. 

Quantum computing holds transformative potential for HCLS by tackling problems intractable for classical computing as well as providing possible, orthogonal interpretations to the vast complexity of biological system~\cite{cellceltric}. Quantum computing may offer new ways of accelerating drug discovery, aiding personalized medicine~\cite{biomarkerWGpaper, Prashant2021}, and scaling optimization problems in healthcare~\cite{Doga2024-vg} such as optimization of treatment plans and managing hospital resources. Indeed, quantum computing offers the potential to unlock new insights and drive significant advancement in HCLS.

Quantum machine learning (QML) has more recently joined the AI/ML and quantum computing fields.  This is an area of research that has grown rapidly over the past twenty years and draws upon the close connections between physical quantum systems and parameterized learning models.  The co-design of algorithms and hardware has led to early successes in the training and deployment of quantum Boltzmann machines \cite{amin2018qbm}, quantum boosting algorithms \cite{neven2012qboost}, and quantum classifiers \cite{farhi2018classification} on quantum annealers. More recently, the availability of gate-based quantum platforms that support trainable unitary operations has led to the development of variational hybrid algorithms that harness quantum models and powerful classical optimization and training workflows \cite{havlivcek2019supervised,peters2021machine}. 

The numerical simulation of circuit behavior is often utilized to study the capabilities of QML models.  Models trained in simulation can be later deployed on hardware. However, building a fully hybrid workflow that can train a QML model using hardware measurements is currently facing a bottleneck due to the number of circuits, circuit depth, and measurements needed in order to implement supervised learning.   

Ensemble methods can potentially alleviate these bottlenecks by training a collection of weak learners: models that are under-parameterized and unable to capture highly complex relationships between features and outputs.  These methods are appealing when mapping applications onto near-term quantum hardware as weak learners can be implemented with smaller, shallower unitary circuits. These methods hold significant promise in HCLS due to the inherent complexity and variability of biological data. As a biological use case, we chose to model immunotherapy response in renal cell carcinoma (RCC) using gene expression data.  While gene expression data has been valuable in modeling response to immunotherapy in cancers such as lung cancer and melanoma, it has  not translated well to RCC despite knowing many of the genes that are directly involved in immunotherapy response.

\subsection{Contributions of this work}
\label{ssec:contributions}
In this work we present results of an empirical study that compares the performance of different quantum ensembling approaches when trained on HCLS data. Our contributions are:
\begin{itemize}
    \item Performance Comparison of different aggregation methods: classical (boosting, bagging and soft-voting) and quantum (boosting via superposition and perturbation)
    \item Overhead comparison for different ensembling workflows:  serial processing using variational learners, parallel processing using cosine learners
    \item Demonstration of efficacy in real-world HCLS data taken from RCC patients treated with immunotherapy \cite{mcdermott2018Clinical}. 

\end{itemize}

\subsection{Related Works}
\label{ssec:related_work}
Aggregation over multiple models and data is  a well-established statistical method \cite{clark1976effects}.  This approach has led to machine learning ensembles built via bagging \cite{breiman1996bagging} and boosting \cite{freund1996experiments}. These approaches have carried forward to more modern ensemble techniques such as XGBoost~\cite{chen2016xgboost}, LightGBM~\cite{ke2017lightgbm}, and CatBoost\cite{prokhorenkova2018catboost}. 

Over the past decade, its been shown that parameterized quantum circuits can be trained as binary classifiers using the qubit readout \cite{havlivcek2019supervised,schuld2020circuit}.  Quantum binary classifiers can be constructed using a single qubit, however the number of qubits used in the model affects the number of classical features that can be embedded into the quantum states \cite{havlivcek2019supervised,schuld2021effect}.
Ensembles of quantum learners have been developed in \cite{schuld2018quantum,abbas2020quantum,Silver_Patel_Tiwari_2022}. Quantum boosting approaches have been developed for adiabatic systems \cite{neven2012qboost} and gate-based systems \cite{macaluso2020quantum}.  For quantum AdaBoost there have been promising results in the analysis of trainability and generalization error \cite{wang2024supervised} but to achieve low error these bounds assume that the number of training samples can grow arbitrarily large.

\section{Methods}
\label{sec:models}
In this work, we are focused on supervised learning methods where algorithms learn patterns and relationships between the underlying data structure and a set of known labels in order to make predictions in unseen or new data. Ensemble learning methods are a subset where rather than train a single classifier, multiple classifiers, whose performance individually may be suboptimal, are aggregated through various methods to yield improved predictions.  Here, we are training ensemble classifiers to predict binary labels.  We leverage standard supervised learning, using datasets of labeled multi-dimensional features $\lbrace (X_i, y_i) \rbrace$. 

We evaluate several constructions of quantum  ensembles: using classical aggregation via soft voting, bagging \cite{breiman1996bagging}, or boosting \cite{freund1996experiments}; and quantum boosting and perturbation\cite{macaluso2020quantum}.  These ensembles are distinguished by the methods used to aggregate predictions from each individual learner, and the feature partitions that each learner trains on.

\subsection{Quantum Ensembles of Quantum Cosine Classifiers}
In this study we apply an implementation of a quantum cosine classifier (QCC) and then an ensemble of those classifiers from \cite{macaluso2020quantum}.
In brief, the quantum cosine classifier uses a swap-test\cite{buhrman2001quantum} to calculate the cosine distance of two sample vectors in a quantum state via interference. Here then for a test sample, it returns the probability of a sample belonging to a class from a single-qubit measurement based on its distance from a randomly selected training sample. This classifier is well suited for ensemble methods because it is a weak classifier with high variance subject to the random selection of the training samples. The QCC as implemented by \cite{macaluso2020quantum} uses one training sample and two features, yielding circuits requiring four qubits.

The quantum ensemble cosine classifier (QEC) uses a quantum circuit to capture the independent quantum trajectories by sampling in superposition from $2^d$ transformations of the training set and then averaging across the predictions, where $d$ is the control register. After this sampling, learning via interference proceeds as defined in the quantum cosine classifier. We refer the reader to \cite{macaluso2020quantum} for further details. We modified the original implementation to enable increased training sample sizes and number of features considered. We tested multiple configurations over a range of parameter values: $d=[1,2,3]$, $n\_train=[2,4]$, $n\_swap=[1,2,4]$, and $n\_feature=[2,4,8]$. These configurations yield circuits whose qubit requirements range from $7-23$.

While the procedure above yields unitary transformations that are uncorrelated in general, a natural question to ask is what the effects of other forms of random sampling on $\mathcal{U}(n)$ are in regards to performance of the ensemble. This opens up essentially all distributions on $\mathcal{U}(n)$ as a possible choice for sampling. We developed a quantum ensemble cosine classifier with random unitaries (QECRU), though given that an exhaustive analysis of all such possible choices is beyond the scope of this work, we focus on the choice that makes the fewest assumptions possible: the uniform distribution on $\mathcal{U}(n)$. We use the function \texttt{scipy.stats.unitary\_group} to generate as many different random unitary operators as required. The implementation follows \cite{mezzadri2007generaterandommatricesclassical}. We tested multiple configurations over a range of parameter values: $d=[1,2,3]$, $n\_train=[2,4]$, $n\_swap=[1,2,4]$, and $n\_feature=[2,4,8]$. These configurations yield circuits whose qubit requirements range from $8-23$ and executed with 8192 shots.

\subsection{Variational Quantum Ensembles}
\label{ssec:variational_ensembles}
Variational quantum classifiers \cite{havlivcek2019supervised,schuld2020circuit} translate classical supervised learning into hybrid workflows. Label predictions $\hat{y}_i$ are made using finite samples sampled from quantum states prepared using parameterized quantum circuits ($\mathcal{U}(x_i,\theta)$). Many approaches found in the literature use the expectation of a fixed observable to predict class labels.  Our approach uses the observe occurrence of the (0/1) bitstrings when qubit 0 is measured in the computational basis. With this approach it is straightforward to extract class probabilities $p(y=0), p(y=1)$ and to train using binary cross entropy loss.

The ``weak learners'' of our variational ensemble are shallow-depth parameterized quantum circuits. From the large design space of  parameterized quantum circuit ansatzes, we use specific design choices and constraints.  We choose amplitude embedding to map a multi-dimensional feature $x_i$ into the $2^n$-dimensional Hilbert space of $n$-qubits. Second, a parameterized single qubit rotation, decomposed as a RZ-RY-RZ gate sequence, is applied to each qubit (3 independent trainable parameters per qubit). If the learner has more than two qubits, this is followed by a layer of CNOT gates applied between qubits $(i, i+1)$. Third, another parameterized rotation is applied to each qubit, followed by classical readout of qubit 0 which is post-processed to make a label prediction. 

A variational learner on $n$ qubits will have $6n$ trainable parameters, and an ensemble of $k$ learners has a total $6nk$ parameters to train. The variational ensembles of classifiers are trained using mini-batch gradient descent with Adam \cite{kingma2014adam} and, using parameter shift rules to evaluate analytic circuit gradients \cite{schuld2019evaluating}.  We optimize hyper-parameters using k-fold cross validation (k=4) and a grid of 90 configurations: three Adam learning rates ($\alpha \in [\num{1e-3},\num{1e-2},\num{1e-1}]$), five batch size ($b \in [1, 2, 4, 8, 16]$), and seven ensemble sizes ($n_{\ell} \in [1,2,3,4,5,6,7]$). We used the Gaussian blobs and use the validation set performance to down-select on optimal ensemble designs, which are re-fit on the full training dataset. In particular, the datasets with overlapping blob centers $p1=p2$ under amplitude encoding, will see all features mapped close to the equator of the Bloch sphere, where the output probabilities $p(y=0) \approx p(y=1) \approx \dfrac{1}{2}$, and with smaller $cluster\_std$ the encoded features will be located in a narrower band around the equator.

\paragraph{Soft voting variational classifier} Every learner generates a predicted class membership based on the probability of observing the single qubit bitstrings $0/1$ in the $|0\rangle$ or $|1\rangle$. The final state label is predicted by aggregating the probabilistic output of sampled outputs using the average of all predictions $\dfrac{1}{K}\sum_k p^{k}(y_i=1)$.  This workflow relies on serial processing of all samples.  For soft-voting, the blob datasets with well-separated cluster centers (e.g. $p1= 0.3, p2=1.$, $p1=0.3,p2=0.5$) and either $cluster\_std=0.3, 0.5$ we observe that all ensemble sizes, all batch sizes trained with learning rates $\alpha=\num{1e-2},\num{1e-1}$ could achieve median validation set accuracy of $70\%$ or higher, however learning rate $\alpha=\num{1e-3}$. For blob datasets with overlapping centers ($p1 = p2 = 1, cluster\_std = 0.5$), or ($p1 = p2 = 0.5, cluster\_std = 0.3$), only a few ensemble configurations could achieve median validation set accuracy above $50\%$.  We choose to retrain ensembles containing up to 4 learners, including a single classifier as a control. With two-dimensional features (1 qubit per learner) these ensembles only require a maximum of 8 qubits to instantiate the ensemble.  However for the RCC data, with 3 qubits required per learner the circuit size grows to 12 qubits. The re-training used learning rates $\alpha = [\num{1e-3},\num{1e-1}]$, and the same configurations (and learning rates) are used to train on the RCC data.

\paragraph{Bagged variational classifier}
A bagged ensemble of [k] classifiers is trained by first partitioning the training data into [k] distinct subsets. The k-th learner is only trained on the k-th data subset.  During the inference stage all learners make a prediction which is aggregated using a weighted mean over each learners' prediction $p(y=1)$. For bagged ensembles, the blob datasets with well-separated cluster centers were easiest to learn -- with learning rates $\alpha=\num{1e-2},\num{1e-1}$ all ensemble configurations could achieve median validation sets accuracy near $100\%$ while for $\alpha=\num{1e-3}$ smaller ensembles and smaller batch sizes performed better. For blob datasets with overlapping cluster centers and smallest $cluster\_std$, three configurations were able to achieve median validation accuracies above $62\%$:  $(b, n_{\ell}) = [(2,3),(4,6),(8,3)]$. We take these three configurations and re-train them on the Gaussian blob datasets using $\alpha = [\num{1e-3},\num{1e-1}]$. The same configurations (and learning rates) are used to train on the RCC data.

\paragraph{Boosted variational classifier}
A boosted ensembles of variational classifiers uses AdaBoost \cite{freund1995desicion,mason1999boosting} or general gradient boosting \cite{friedman2001greedy}. AdaBoost iteratively updates the weight (importance) of individual samples in the training set. Training samples are initially equal weighted and a weak classifier is trained on a random subset (drawn without replacement).  The weak learner's error on the entire training set is used to update the sample weights, then a new learner is trained on a different training subset using the updated weights. For blob datasets with overlapping cluster centers, AdaBoost was not able to train any ensemble that had a median validation accuracy above $50\%$ (random guessing).  Instead we retain the top three individual configurations: $(b, n_{\ell}) = [(2,3),(4,6),(8,3)]$ and re-train this on the Gaussian blob datasets using $\alpha = [\num{1e-3},\num{1e-1}]$, and use the same configurations and learning rates to train on RCC data features.

\subsection{Classical Ensembles}
\label{ssec:classical_ensembles}
In this study, classical ensembles are represented by random forests (RF), a popular ensemble method for both classification and regression problems that is an extension of the bagging technique with added randomness to enhance diversity among decision trees~\cite{breiman2001random}. In brief, random forests utilize decision trees from random subsets of samples and features to induce independence between trees.  Voting is traditionally a `majority rule'. RF is robust and highly adept at handling high-dimensional datasets with complex interactions with demonstrated advantages on tabular data with respect to newer deep learning methods~\cite{grinsztajn2022tree}. They also provide measures of variable importance. Yet, RF can be challenged by small datasets and highly correlated features. We applied RF classifiers using \textit{scikit-learn} (\textit{v1.6.1}) with parameter optimization using the $RandomizedSearchCV$ function from \textit{scikit-learn} testing the following parameter ranges: $n\_estimators = [100..1000]$  by step 100, $max\_depth = [5..20]$, $min\_samples\_split = [2..10]$, $min\_samples\_leaf = [1..5]$, and $max\_features = [sqrt, log2]$. For a given train/test split, this randomized parameter search was performed on the training set and the identified best parameters are then applied on the test set.

\subsection{Datasets and Feature Selection}
\label{ssec:data_features}
\begin{figure}
    \includegraphics[width=\columnwidth]{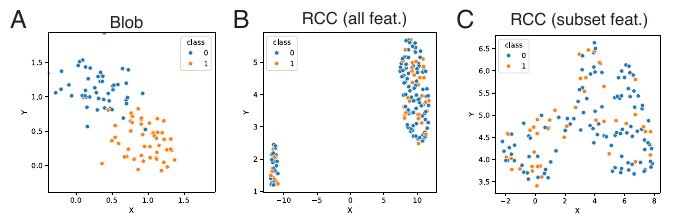}
    \centering
    \caption{UMAP projections of study datasets. A) Example of one tested configuration of Gaussian blobs. In this configuration, 100 samples are evenly drawn from two classes, class 0 and 1.  Class 0 center coordinates are $(0.3,1.0)$ with $\sigma=0.3$, class 1 coordinates are $(1.0,0.3)$ with $\sigma=0.3$. B-C) McDermott RCC datasets are projected using all gene features (B) and 8 gene features selected from literature (C).}
    \label{fig:datasets}
\end{figure}

\paragraph{Gaussian Blobs} The Gaussian blob benchmark is a synthetic dataset that was generated in \textit{scikit-learn} (\textit{v1.6.1}). Eighteen different configurations are used to create sets of two-dimensional features sorted into two classes over all combinations of the following parameters: $cluster\_std=[0.3,0.5]$, $p1=[0.3, 0.5, 1.0]$, and $p2=[0.3, 0.5, 1.0]$. $p1$ and $p2$ represent the x and y coordinates of the centers of class 1 ($p1$, $p2$) and class 2 ($p2$, $p1$) blobs.   The features are generated in the domain $[-0.85, 2.55] \times [-0.85, 2.55]$ and are rescaled to [0, 1] using min-max scaling without standardization.  The blob datasets contain 100 labeled samples and 10 unique 80/20 train/test splits are generated (Fig. \ref{fig:datasets}).  
For hyperparameter tuning, the training data is split into 5 folds for the random forest and 4 fold for the variational ensembles. 

\paragraph{Renal Cell Cancer (RCC)}
From the renal cell cancer cell dataset~\cite{mcdermott2018Clinical} we use DESeq2~\cite{love2014moderated} to normalize RNA-seq mRNA counts for 150 patient samples. We also applied variance stabilizing transformation (VST) from DESSeq2 to ensure constant variance across the range of mean values seen per sample. For analyses, we use both the entire gene feature space as well as 8 hand-selected genes (\textit{CD8A}, \textit{CXCL9}, \textit{CXCL13}, \textit{IFNG}, \textit{CD274}, \textit{PDCD1}, \textit{VHL}, \textit{GZMK}), which are known to associate with immunotherapy response \cite{Litchfield2021Cell} \cite{Zhen2025NatCom}.These features are rescaled from the original domains of (ppm) to $[0, 1]$ using a min-max scaling without standardization. For dimensional reduction, principal component analysis (PCA) was performed and the first $f$ components were used as features where $f$ was an experimental parameter. The RCC dataset's 150 samples were divided into 10 train/test splits (80/20).  We visualize the RCC datasets using two-dimensional features extracted using the uniform manifold approximation and projection (UMAP) algorithm \cite{mcinnes2018umap}.

\begin{figure*}
\centering
    \includegraphics[width=0.9\textwidth]{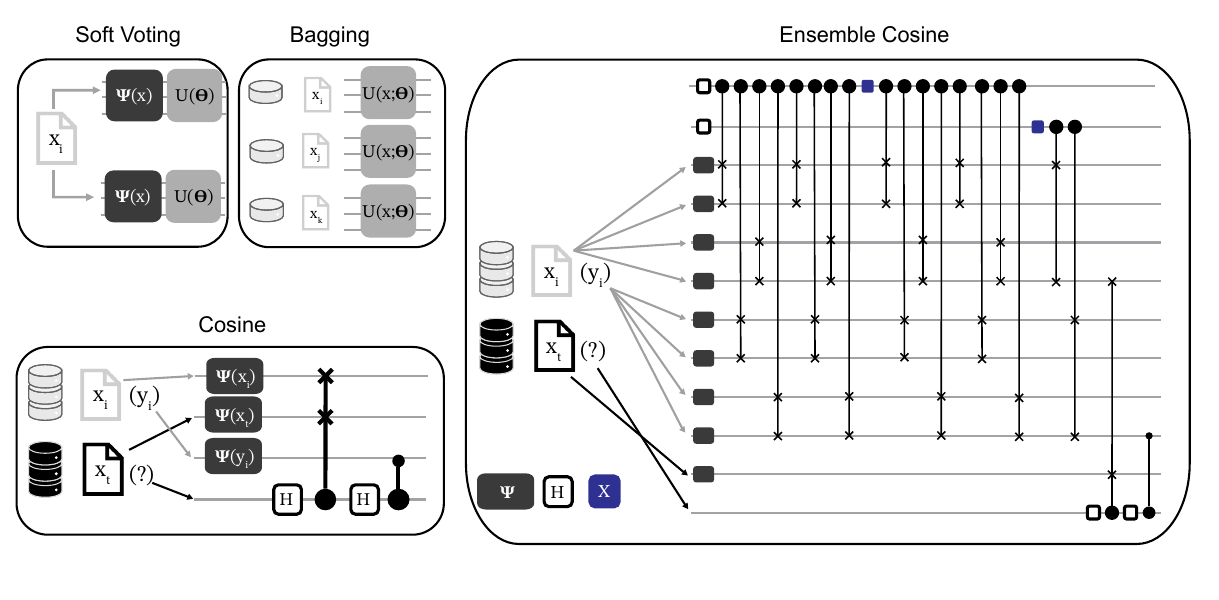}
    \caption{Comparison of quantum ensemble constructions.  Two variational ensembles (Soft Voting and Bagging), the cosine classifier, and the quantum ensemble cosine classifier are shown. In the variational ensembles each learner predicts the label of one encoded feature vector $x_i$ at a time using a parameterized unitary. In the Soft Voting panel we show the unitary construction of the variational classifier-- which we condense into $\mathcal{U}(x;\theta)$ in the remaining panels.  Cosine classifiers predict an unknown label using multiple encoded feature vectors $x_i, x_t$ and also the known label information $y_i$.  The quantum ensemble cosine classifier has ancilla qubits for the control register (top two qubits here). In this is example four unique training samples are used to predict a single test sample. The non-variational ensembles do not have trainable parameters.  During variational training samples are drawn from either the entire training dataset and used to train all learners (Soft Voting), or the dataset is partitioned into non-overlapping subsets and each learner only has access to a specific subset (Bagging).  Boosting (not shown) uses random weighted subsets.}
\end{figure*}
\section{Results}
\label{sec:results}

\subsection{Classifier Performance}
\label{ssec:metrics}

We assess the performance of our models using the true positive (TP), true negative (TN), false positive (FP) and false negative (FN) to compute the accuracy ($\dfrac{(TP+TN)}{(TP+TN+FP+FN)}$); weighted F$_1$ score from \textit{scikit-learn}; and the Brier score $S = \langle (p_i - y_i)^2\rangle$. The probability of outcome $y_i$ used in the Brier score is defined by the number of shots observed in the 0 or 1 bitstring. For accuracy and weighted F$_1$ metrics, higher scores indicate better predictions. For the Brier score, lower scores indicate better predictions.

\begin{figure*}
    \includegraphics[width=0.9\textwidth]{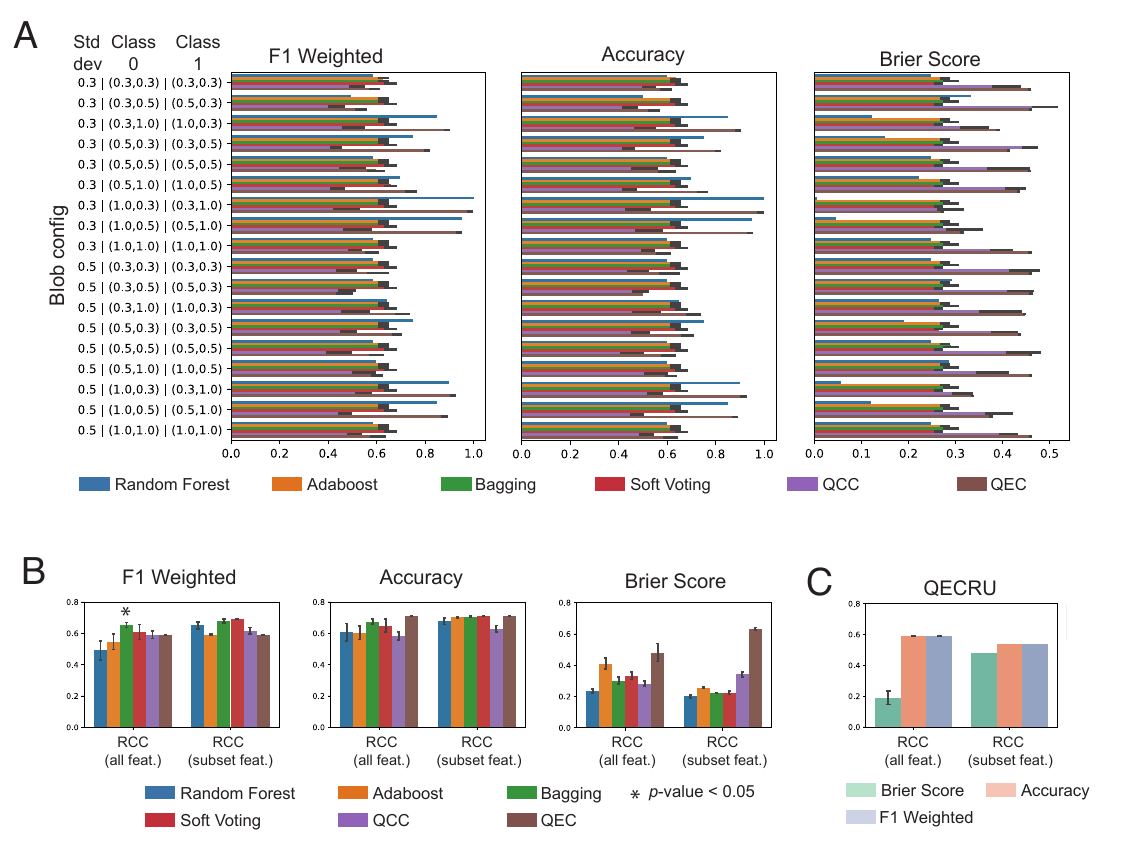}
    \centering
    \caption{Comparison of ensemble classifier performance. Mean and standard error of each metric is calculated over splits for blob (A) and RCC (B) datasets for the random forest, AdaBoost variational ensemble, bagging variational ensemble, soft vote variational ensemble,  quantum cosine classifier (QCC), quantum ensemble of quantum cosine classifiers (QEC), and QEC with random unitaries (QECRU).  A) Performance per blob configuration for a given classifier configuration over 10 splits. Blob configurations are indicated along the y-axis with specified standard deviation and \textit{x,y} coordinates for centers of class 0 and 1. For a given blob configuration and classifier type (random forest (blue), quantum cosine (orange), and quantum ensemble (green)), we plot the classifier configuration that achieves the maximum mean performance. B) Performance of classifiers on RCC datasets over 10 splits, including both full feature space and 8 feature subset. Asterisks indicates significant improved over random forest calculated by $t$-test. C) Performance of QECRU over five splits.
    }
    \label{figure:model_perf}
\end{figure*}

For each model described in Section \ref{sec:models}, we compared the trained ensemble performance across the 18 blob configurations (Fig. \ref{figure:model_perf}A). For each classifier type, we identify the configuration with the maximal mean performance over all 10 splits. We observe that RF and QEC perform similarly when considering the accuracy and F$_1$, and both outperform the QCC. QCC reached an F$_1$ or accuracy $\approx0.5$ despite blob configurations that were well separated, as may be expected from being a single weak learner. As compared to the variational bagging, AdaBoost, and soft voting classifiers, these quantum ensembles on average out perform the RF, with a few exceptions where the RF and QEC significantly outperforms the variational methods. We note that RF consistently achieves a lower Brier score than all quantum approaches.

When we compared classifier performance on the RCC dataset (Fig. \ref{figure:model_perf}B,C and Table~\ref{tab:model_comparison}), we found that the quantum classifiers showed similar to slightly improved F$_1$ and accuracy scores as compared to the RF, though the margin does not rise to significance, with the exception of the bagging variational classifier whose F$_1$ score was significantly higher than RF ($p$ = 0.018 by $t$-test) when testing on the full RCC dataset. Among the quantum methods, when we identified the maximal performance for any split, we see the bagging classifier also reached the highest F$_1$ of 0.81 (Table ~\ref{tab:model_comparison}).
When testing on the RCC dataset with 8 selected gene features, all methods had comparable performance.

Given the protracted execution times to simulate the QECRU, we report its performance over 5 splits of the data (Fig. \ref{figure:model_perf}C), and find it to perform at par with the QEC, albeit with an improved Brier score. We did also find that when given the entire dataset, the random forest predicted only a single class for 9 of the 10 splits. This undesired behavior was absent when using the QCC, and appeared to lesser degrees for the QEC and QECRU with 4 of 10 and 2 of 5 splits predicted as a single class, respectively. For those splits where the classifiers were not pathologic in their predictions, we observe similar mean performance values for the RF ($F_1=0.52$), quantum cosine ($F_1=0.61$), quantum ensemble ($F_1=0.58$), and quantum ensemble with random unitaries($F_1=0.60$).

\begin{figure*}
    \includegraphics[width=0.7\textwidth]{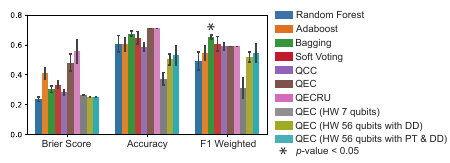}
    \centering
    \caption{Comparing simulation and quantum hardware QEC performance.  Performance of the QEC executed on quantum hardware, $ibm\_kyiv$ with 7 (purple) and 56 (brown) qubits are shown. These classifiers were run over five splits of the RCC data with PCA embeddings with 2 and 32 principle components as features for the 7 and 56 qubit experiments, respectively, using either Pauli twirling (PT) or PT and dynamical decoupling (DD) error mitigation. Classifiers shown in Figure~\ref{figure:model_perf} are included for ease of comparison. Mean and standard error of each metric is calculated over splits for RCC datasets. Asterisks indicates significant improvement over random forest calculated by $t$-test.
    }
    \label{figure:model_perf_HW}
\end{figure*}

Further, we performed a preliminary set of experiments on a 127 qubit quantum device, $ibm\_kyiv$, using the QEC on the RCC dataset based on well performing configurations in simulation over five splits (Fig.~\ref{figure:model_perf_HW}). These experiments were performed using the Qiskit SamplerV2 primitive with 8192 shots and Pauli twirling (PT) and dynamical decoupling (DD) with \textit{XY4} gate sequence error mitigation active. The first experiment using 7 qubits was configured as: $d=1$, $n\_train=2$, $n\_swap=1$, and $n\_feature=2$.  This yields a circuit with overall depth and 2-qubit depth of 100 and 20, respectively.  This configuration underperformed the other ensemble classifiers and the random forest. We then ran a larger 56 qubit circuit first using only PT and then with both PT and DD  with the following configuration:  $d=2$, $n\_train=8$, $n\_swap=1$, and $n\_feature=32$. The transpiled depth was significantly deeper with overall depth and 2-qubit depth of 853 and 201, respectively. With only PT, the configuration underperformed QEC in simulation. With PT and DD error mitigation, performance improved and reach a similar weighted F$_1$ to QCC, QEC, and QECRU, which is slightly improved to the random forest, while reaching a significantly lower Brier score than QEC and QECRU.

\begin{table}
    \centering
    \begin{tabular}{|c|c|c|c|c|c|}
    \hline
        \textbf{Model} & \textbf{Features} & \textbf{Qubits} & \textbf{Depth} & \textbf{2Q depth} & \textbf{F$_1$}\\
    \hline
    \multicolumn{6}{|c|}{Single Classifiers}      \\
    \hline
        Cosine & 2 & 4 & 64 & 11 & 0.77 \\
        Variational & 8 & 3 & 59 & 14 &  0.78 \\
    \hline
    \multicolumn{6}{|c|}{Ensemble Models}      \\
    \hline
        Cosine & 4 & 16 & 132 & 27 & 0.72 \\
        Random Unitary & 8 & 8 & 143 & 30 & 0.66\\
        Soft Vote & 4  & 10  & 22  & 12 & 0.78 \\
        Bagging & 8  & 9 & 57  & 14 & 0.81 \\
        AdaBoost  & 2  & 6 & 6  & 0 & 0.68 \\
    \hline
    \end{tabular}
    \vspace{.1cm}
    \caption{Comparing best performing ensembles on an individual RCC test dataset split using a PCA feature embedding from the full feature space, and reporting on circuit width, circuit depth including measurement, feature size, and weighted F$_1$ score.}
    \label{tab:model_comparison}
\end{table}

\section{Analysis}
\subsection{Comparison to classical baseline}
\label{ssec:classical_baseline}
The performance results reported in Figure \ref{figure:model_perf}, Figure \ref{figure:model_perf_HW}, and Table \ref{tab:model_comparison} shows a comparison between the models.  In this Section we analyze these performance metrics focusing on the simulation results where more extensive experimentation was performed to further assess the effects of: qubit overhead, embedding structure, and feature structure.

We found there were slight differences in performance between the variational and cosine quantum ensembles. 
In synthetic data, the variational approaches slightly outperformed the RF and cosine classifiers for many of the blob configurations. However for blob configurations whose centers were further apart it is notable that the RF and QEC were able to reach much higher levels of performance than the other approaches. It would be expected that these configurations where the classes are better separated would see higher predictive performance from all methods.  It may be the variational methods are relatively over-parameterized for these simplistic two feature datasets. In contrast for the significantly more complex RCC dataset, we find the variational methods performing well with the bagging classifier significantly improving on the RF. The other quantum classifiers were doing well compared to RF with a modest, though insignificant, increase of the F$_1$ or accuracy by $\approx10\%$. While the performance of QEC and QECRU was not significantly improved over the variational and RF classifiers, it was able to achieve its promising performance only using anywhere from 2-4 training samples. This suggests the potential for significant applications to cases where the amount of data is highly constrained and methods, such as QEC, are still able to effectively learn.

Though the performance of the variational ensembles were comparable to the quantum cosine ensembles, the variational ensemble sizes were limited to fewer than 10 learners based on the expensive hyperparameter grid search. As a result the variational ensembles fully trained on the RCC data used more samples to train each learner:  soft voting used all training samples per learner, bagging used $|X|/\ell$ samples per learner, and boosting also used $|X|/\ell$ samples per learner.  Variational ensembles could be trained using fewer samples per learner if the number of learners in the ensemble was increased to $\ell > 25$.  We evaluated the performance of training an AdaBoost ensemble on the top 8 principal components in the RCC dataset.  These results are not included in Fig. \ref{figure:model_perf} but the best performance on the test set was comparable to the best AdaBoost performance reported in Table \ref{tab:model_comparison}. Therefore both variational and cosine quantum ensembles share this value of being able to learn using fewer samples with broad applications to data-constrained problems.

\subsection{Overhead}
\label{ssec:overhead}
To assess the utility of our approaches, we consider the overhead and scaling of each model (qubit overhead, gate depth). The ensemble cosine classifier and random perturbation of trajectory model utilizes a parallelization in feature processing. This enables $2^d$ transformations of the input state with linear cost in circuit. These classifiers' qubit overhead scales logarithmically with the feature size and linearly with the number of training sample size and number of control registers.  Transpiled circuit depth though grows rapidly with the register size, number of swaps, and training size, particular with the prevalence of long distance $CNOT$ gates. With 56 qubits acting on 32 eight features, 2 control registers, 8 training samples, and 1 random swap, the circuit has an overall and 2-qubit depth of 853 and 201, respectively (Table~\ref{tab:model_comparison}). It is possible that some of these circuit depth challenges can be mitigated with circuit knitting techniques. 

The  feature scaling of the base variational classifier is the same in qubit overhead, using $\log({f})$ qubits to encode $f$ classical features, but the dependence on supervised training makes them less efficient to scale up.  For the bagged and boosted ensembles, the learners are trained one at a time on disjoint subsets of the training data and ensemble predictions are aggregated only during inference. This made the numerical simulation of bagged and boosted ensembles quite fast as each learner was constructed and trained one at a time -- for the largest feature size of $8$ this required training a circuit with at most $3$ qubits and $18$ trainable parameters. The bagged ensemble training could be easily distributed, but not the boosted ensemble training due to the need to adaptively weight the training samples.   

On the other hand, the soft-voting ensemble was the slowest to train and incurred the highest simulation overhead. These ensembles used the highest amount of memory during training and the gradient update step was a major bottleneck and this is wholly due to how the soft-voting learners were updated during training. 
 
Implementing the gradient update for each learner and ensemble could be done in multiple ways.  For the soft-voting ensemble, the ensemble predictions are needed to evaluate the loss at each step of gradient descent. We implemented the soft-voting ensemble training without distributing the execution of each individual learner and generated the ensemble predictions by sampling from the state prepared in a single circuit of $n \times \ell$ qubits: $\Psi = \mathcal{U}_{0} \otimes \mathcal{U}_{1} \otimes \dots \mathcal{U}_{\ell}$. 

The trained variational ensembles were not deployed on hardware, and this poses an open question about the number of quantum resources each ensemble needs.  The aggregation of the learner predictions is implemented classically as a post-processing step. Thus to generate predictions from an ensemble of $\ell$ learners we can execute each learner serially using a dedicated $n$ qubit state preparation and measurement circuit for each.  However this does not seem like an efficient use of near-term quantum processors which offer far more qubits than the number used per learner in this study.  The second approach would be to simultaneously prepare all $\ell$ learner states using disjoint hardware qubit subsets.  However, as the register size increases the sampling overhead will also increase.  Additionally, when executed on hardware the assumption that each learner will remain independent and unperturbed by the gates applied to other learners is dependent on the presence of correlated noise.

\section{Conclusions}
HCLS applications have many opportunities for hybrid ML models, and it remains an open question how to optimally incorporate new processing modalities and to identify which problems a given algorithm will be most adept.  The ability to learn from fewer training examples is highly sought after in HCLS applications where biological samples are typically difficult to acquire and problems are often under-determined. This sort of challenge is exemplified when new phenotypes of interest emerge. For example, a new viral strain is discovered and an ML model to predict susceptibility would either rely on transfer learning from data of known strains or be required to learn from the limited samples in this new exposure. Either scenario is a challenge for classical machine learning in HCLS. The comparable-to-improved performance achieved by the QEC on few training samples in the renal cell cancer dataset suggests their applicability in these data-constrained HCLS applications. 

Quantum ensemble models are a potential path to utility scale QML applications: replacing quantum deep learning approaches with quantum methods closer to random forests or using random subspace sampling to extract data to train weak learners. We have presented results that show how quantum ensemble models can to learn from fewer training examples. The potential to harness quantum superposition as an inherent parallelization is a particularly attractive feature of these methods. For the cosine classifiers, the input state moves through multiple quantum trajectories in superposition as a function of $d$ control registers that generates $2^d$ transformations. By averaging over these different trajectories, the quantum ensemble cosine classifier needs just a single measurement to obtain a prediction. This enables exploring a vast landscape of learner with relatively review qubits, though with the need to balance circuit depth.

Utility-scale quantum computing typically requires large qubit registers (50+) to reach complexity unable to simulated classically and even in the presence of hardware noise, noisy unitary operations, there must be usable signal that can be extracted from measurements.  Comparing the different models we see a tradeoff between models that leverage superposition but requires ancilla qubits and long range connections, or variational models that require serial processing and parameterized gates. Our initial experiments deploying a 56 qubit QEC on an Eagle IBM QPU (Fig.~\ref{figure:model_perf_HW}), demonstrates that with minimal error mitigation and with only 8 training samples, the QEC reached Brier scores comparable to random forests and improved upon the performance of smaller QECs in noise-free simulation. The QEC model can scale up to larger and more complex circuits by increasing the number of control registers, training samples, and features. To build towards utility-scale demonstrations, more complex circuits, additional error mitigation strategies, and execution on more advanced QPUs with greatly reduced noise, such as IBM Heron, are needed.  

With the promising results of these initial experiments, we believe that this opens the way for the use of quantum ensemble approaches to model biomarkers of the immunotherapy response. The complex interplay between gene expression networks appears to be more easily identified in small datasets using this approach, as many other standard machine learning methods failed to reach the same precision. These methods also lay a framework for combining different data modalities such as whole exome sequencing-based measurements of copy number changes and tumor mutational burden alongside gene expression. Capturing the multi-modal interaction of complex biological features is of great interest for clinical trial design and the identification of personalized medicine approaches.

Quantum ensemble classifiers, as a subfield of quantum machine learning, represent an important frontier in computational technology. These classifiers combine the strengths of classical ensemble learning and the strength of quantum computing. These methods have potential to learn more effectively or generalize better than classical ML. For under-determined or data-constrained problems as is often encountered in  HCLS applications, quantum ensembles can be a critical tool in advancing HCLS research.

\section{Acknowledgments}
This work was supported in part by the U.S. Department of Energy (DOE), Office of Science, Office of Advanced Scientific Computing Research (ASCR), the Accelerated Research in Quantum Computing (ARQC) program under FWP ERKJ450.

\bibliographystyle{unsrt}  
\bibliography{refs}
\end{document}